\documentclass[11pt,a4paper]{article}
\usepackage{authblk}
\usepackage[hyperref]{emnlp2018}
\usepackage{times}
\usepackage{latexsym}

\usepackage[utf8]{inputenc} 
\usepackage[T1]{fontenc}    
\usepackage{nicefrac}       
\usepackage{microtype}      
\usepackage{url}
\usepackage{amsfonts}
\usepackage{amsmath}
\usepackage{amssymb}
\usepackage{amsthm}
\usepackage{mathrsfs}
\usepackage{tikz}
\usepackage{graphicx}
\usepackage{algpseudocode}
\usepackage{verbatim}
\usepackage{xcolor}
\usepackage{booktabs}
\usepackage{multirow}
\usepackage{todonotes}
\usepackage{footnote}
\makesavenoteenv{tabular}
\usepackage[font=small]{caption}
\usepackage{mwe}
\usepackage{lipsum}
\usepackage{titlesec}
\usepackage{subfig}
\usepackage[linesnumbered,ruled]{algorithm2e}
\aclfinalcopy 

\renewcommand{\tt}[1]{\fontfamily{cmtt}\selectfont #1}

\newcommand{\eos}{$\langle \text{eos} \rangle$}
\newcommand{\eop}{$\langle \text{eop} \rangle$}
\newcommand{\ie}{\textit{i.e.}}
\newcommand{\eg}{\textit{e.g.}}

\newcommand{\seqtoseq}{\textit{seq2seq}}
\newcommand{\treedec}{TrDec}

\title{A Tree-based Decoder for Neural Machine Translation}

\author[1]{Xinyi Wang}
\author[1,2]{Hieu Pham}
\author[1]{Pengcheng Yin}
\author[1]{Graham Neubig}
\affil[ ]{\texttt{\{xinyiw1,hyhieu,pcyin,gneubig\}@cs.cmu.edu}}
\affil[1]{Language Technology Institute, Carnegie Mellon University, Pittsburgh, PA 15213}
\affil[2]{Google Brain, Mountain View, CA 94043}

\date{}

\begin{document}
\maketitle
\begin{abstract}
  Recent advances in Neural Machine Translation (NMT) show that adding syntactic information to NMT systems can improve the quality of their translations. Most existing work utilizes some specific types of linguistically-inspired tree structures, like constituency and dependency parse trees. This is often done via a standard RNN decoder that operates on a linearized target tree structure. 
However, it is an open question of what specific linguistic formalism, if any, is the best structural representation for NMT. In this paper, we (1) propose an NMT model that can naturally generate the topology of an arbitrary tree structure on the target side, and (2) experiment with various target tree structures.
%
Our experiments show the surprising result that our model delivers the best improvements with balanced binary trees constructed \textit{without} any linguistic knowledge; this model outperforms standard \seqtoseq~models by up to 2.1 BLEU points, and other methods for incorporating target-side syntax by up to 0.7 BLEU.%
\footnote{Our code is available at \href{https://github.com/cindyxinyiwang/TrDec_pytorch}{https://github.com/cindyxinyiwang/TrDec\_pytorch}.}

%
\end{abstract}


\section{\label{sec:intro}Introduction}



Most NMT methods use sequence-to-sequence (\seqtoseq) models, taking in a \emph{sequence} of source words and generating a \emph{sequence} of target words~\citep{kalchbrenner,seq2seq,attention}. While \seqtoseq~models can \emph{implicitly} discover syntactic properties of the source language~\cite{shi}, they do not explicitly model and leverage such information.
Motivated by the success of adding syntactic information to Statistical Machine Translation (SMT)~\cite{GalleyHKM04,MenezesQ07,GalleyGKMDWT06}, recent works have established that explicitly leveraging syntactic information can improve NMT quality, either through syntactic encoders~\citep{source_syntax_nmt,Tree2SeqEriguchiHT16}, multi-task learning objectives \cite{syntax_aware_nmt,RNNG_EriguchiTC17}, or direct addition of syntactic tokens to the target sequence \cite{CCG17,str2lin_tree}. 
However, these syntax-aware models only employ the standard decoding process of \seqtoseq~models, \ie~generating one target word at a time.
One exception is~\citet{seq2dep}, which utilizes two RNNs for generating target dependency trees. Nevertheless,~\citet{seq2dep} is specifically designed for dependency tree structures and is not trivially applicable to other varieties of trees such as phrase-structure trees, which have been used more widely in other works on syntax-based machine translation.
One potential reason for the dearth of work on syntactic decoders is that such parse tree structures are not friendly to recurrent neural networks~(RNNs).

In this paper, we propose \emph{\treedec}, a method for incorporating tree structures in NMT. \treedec~simultaneously generates a target-side tree topology and a translation, using the partially-generated tree to guide the translation process~(\autoref{sec:generation}).
\treedec~employs two RNNs: a \textit{rule RNN,} which tracks the topology of the tree based on rules defined by a Context Free Grammar (CFG), and a \textit{word RNN,} which tracks words at the leaves of the tree~(\autoref{sec:model}). 
This model is similar to neural models of tree-structured data from syntactic and semantic parsing~\citep{dyer-EtAl:2016:N16-1,seq2tree,YinN17}, but with the addition of the word RNN, which is especially important for MT where fluency of transitions over the words is critical.

\begin{figure*}[htb!]
  \centering
  \includegraphics[width=0.26\textwidth]{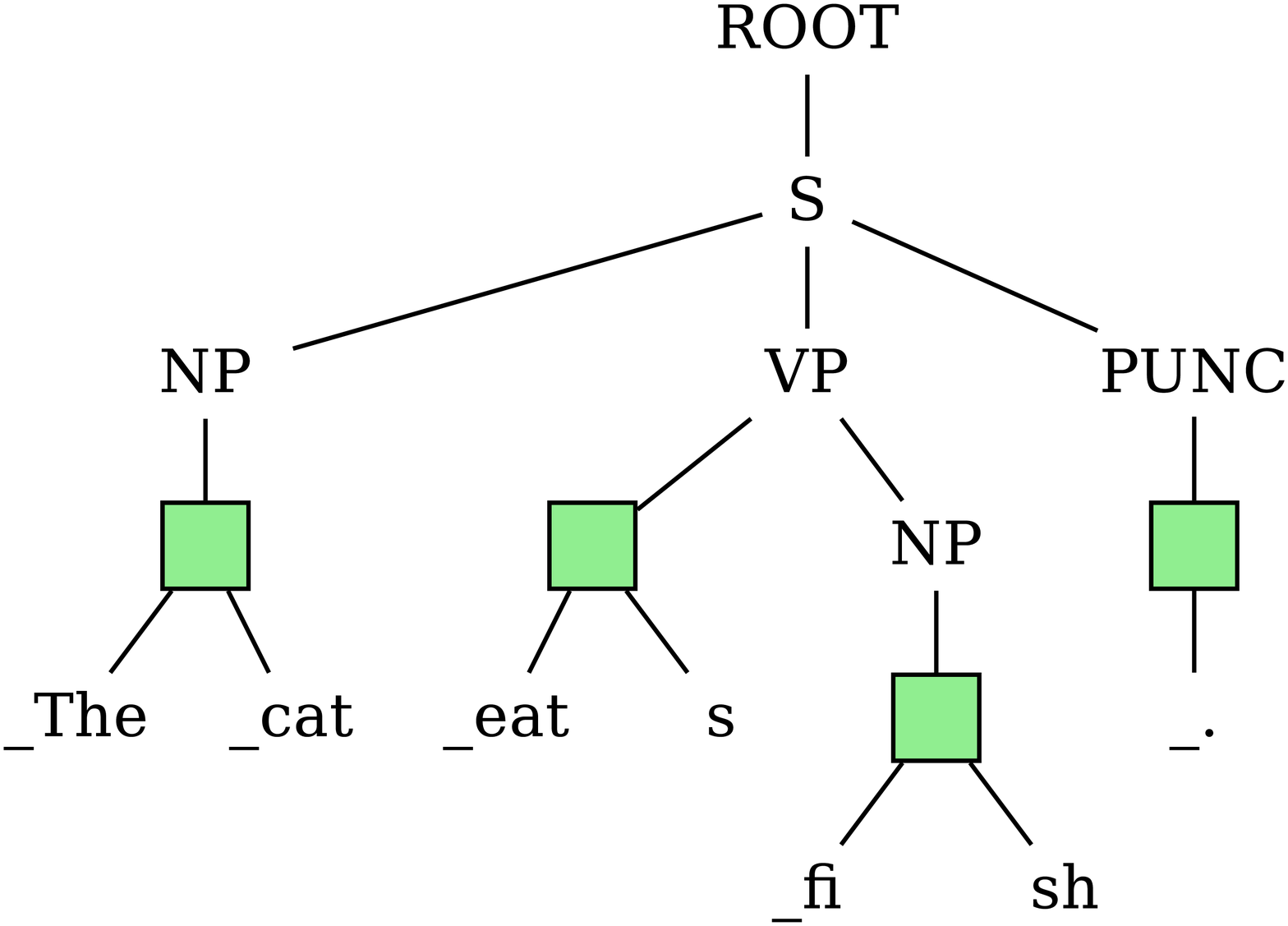}
  ~~
  \includegraphics[height=0.71\textwidth,angle=90]{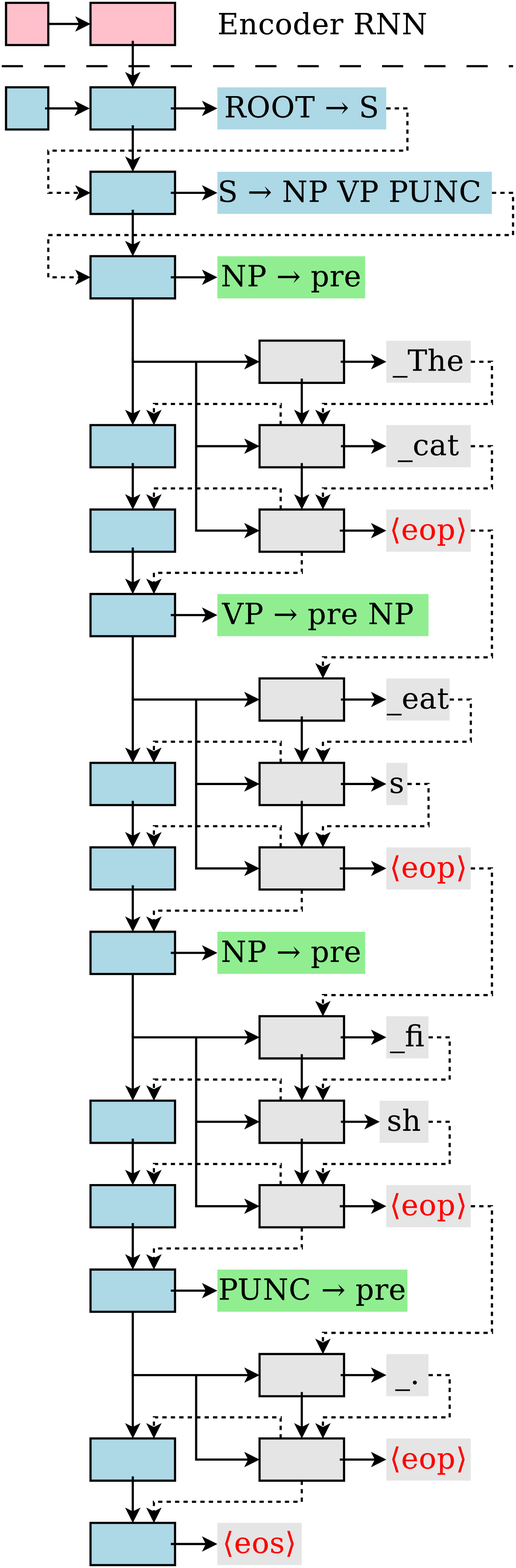}
  \caption{\label{fig:gen_example}An example generation process of \treedec.~\textit{Left:} A target parse tree. The green squares represent preterminal nodes.~\textit{Right:} How our RNNs generate the parse tree on the left. The blue cells represent the activities of the rule RNN, while the grey cells represent the activities of the word RNN. \eop~and \eos~are the~\textit{end-of-phrase} and~\textit{end-of-sentence} tokens. Best viewed in color.}
\end{figure*}

\treedec~can generate any tree structure that can be represented by a CFG. These structures include linguistically-motivated \emph{syntactic} tree representations, \eg~constituent parse trees, as well as \emph{syntax-free} tree representations, \eg~balanced binary trees (\autoref{sec:tree}). This flexibility of \treedec~allows us to compare and contrast different structural representations for NMT.

In our experiments~(\autoref{sec:exp}), we evaluate \treedec~using both syntax-driven and syntax-free tree representations. We benchmark \treedec~on three tasks: Japanese-English and German-English translation with medium-sized datasets, and Oromo-English translation with an extremely small dataset.
Our findings are surprising -- \treedec~performs well, but it performs the best with balanced binary trees constructed \emph{without} any linguistic guidance.

\section{\label{sec:generation}Generation Process}
\treedec~simultaneously generates the target sequence and its corresponding tree structure. We first discuss the high-level generation process using an example, before describing the prediction model (\autoref{sec:model}) and the types of trees used by \treedec~(\autoref{sec:tree}).



\autoref{fig:gen_example} illustrates the generation process of the sentence ``{\tt \_The \_cat \_eat s \_fi sh \_.}'', where the sentence is split into subword units, delimited by the underscore ``{\tt\_}''~\citep{bpe}.
The example uses a syntactic parse tree as the intermediate tree representation, but the process of generating with other tree representations, \eg~syntax-free trees, follows the same procedure.

Trees used in \treedec~have two types of nodes: \textit{terminal nodes}, \ie~the leaf nodes that represent subword units; and \textit{nonterminal nodes}, \ie~the non-leaf nodes that represent a span of subwords. 
Additionally, we define a \textit{preterminal node} to be a nonterminal node whose children are all terminal nodes. In~\autoref{fig:gen_example}~\textit{Left}, the green squares represent preterminal nodes.



\treedec~generates a tree in a top-down, left-to-right order. 
The generation process is guided by a CFG over target trees, which is constructed by taking all production rules extracted from the trees of all sentences in the training corpus.
Specifically, a \textit{rule RNN} first generates the top of the tree structure, and continues until a preterminal is reached. Then, a \textit{word RNN} fills out the words under the preterminal. The model switches back to the \textit{rule RNN} after the \textit{word RNN} finishes.
This process is illustrated in~\autoref{fig:gen_example}~\textit{Right}. Details are as follows:

\noindent \textbf{Step 1.}~The source sentence is encoded by a sequential RNN encoder, producing the hidden states.

\noindent \textbf{Step 2.}~The generation starts with a derivation tree with only a {\tt Root} node. A \textit{rule RNN,} initialized by the last encoder hidden state computes the probability distribution over all CFG rules whose left hand side (LHS) is {\tt Root}, and selects a rule to apply to the derivation. In our example, the rule RNN selects {\tt ROOT} $\mapsto$ {\tt S}.

\noindent \textbf{Step 3.}~The rule RNN applies production rules to the derivation in a top-down, left-to-right order, expanding the current \emph{opening nonterminal} using a CFG rule whose LHS is the opening nonterminal. In the next two steps, \treedec~applies the rules {\tt S} $\mapsto$ {\tt NP} {\tt VP} {\tt PUNC} and {\tt NP} $\mapsto$ {\tt pre} to the opening nonterminals {\tt S} and {\tt NP}, respectively. Note that after these two steps a preterminal node {\tt pre} is created.

\noindent \textbf{Step 4a.}~Upon seeing a preterminal node as the current opening nonterminal, \treedec~switches to using a \textit{word RNN}, initialized by the last state of the encoder, to populate this empty preterminal with phrase tokens, similar to a \seqtoseq~decoder.
For example the subword units {\tt \_The} and {\tt \_cat} are generated by the word RNN, ending with a special \textit{end-of-phrase} token, \ie~\eop.

\noindent \textbf{Step 4b.}~While the word RNN generates subword units, the rule RNN also updates its own hidden states, as illustrated by the blue cells in~\autoref{fig:gen_example}~\textit{Right}.

\noindent \textbf{Step 5.}~After the word RNN generates \eop, \treedec~switches back to the rule RNN to continue generating the derivation from where the tree left off. 
In our example, this next stage is the opening nonterminal node {\tt VP}. From here, \treedec~chooses the rule {\tt VP} $\mapsto$ {\tt pre} {\tt NP}.

\treedec~repeats the process above, intermingling the rule RNN and the word RNN as described, and halts when the rule RNN generates the \textit{end-of-sentence} token \eos, completing the derivation.


\section{\label{sec:model}Model}
We now describe the computations during the generation process discussed in~\autoref{sec:generation}. At first, a source sentence $\mathbf{x}$, which is split into subwords, is encoded using a standard bi-directional Long Short-Term Memory (LSTM) network~\citep{lstm}. This bi-directional LSTM outputs a set of hidden states, which \treedec~will reference using an attention function~\citep{attention}.

As discussed, \treedec~uses two RNNs to generate a target parse tree. In our work, both of these RNNs use LSTMs, but with different parameters.

\paragraph{Rule RNN.} At any time step $t$ in the rule RNN, there are two possible actions. If at the previous time step $t-1$, \treedec~generated a CFG rule, then the state $\mathbf{s}^{\text{tree}}_t$ is computed by:
\begin{equation*}
  \label{eqn:rule_rnn}
  \begin{aligned}
  \small
  \mathbf{s}^{\text{tree}}_t = \text{LSTM}([\mathbf{y}^{\text{CFG}}_{t-1}; \mathbf{c}_{t-1}; \mathbf{s}^{\text{tree}}_p; \mathbf{s}^{\text{word}}_t], \mathbf{s}^{\text{tree}}_{t-1})
  \end{aligned}
\end{equation*}
where $\mathbf{y}^{\text{CFG}}_{t-1}$ is the embedding of the CFG rule at time step $t-1$; $\mathbf{c}_{t-1}$ is the context vector computed by attention at $\mathbf{s}^{\text{tree}}_{t-1}$, \ie~input feeding~\citep{dot_prod_attention}; $\mathbf{s}^{\text{tree}}_p$ is the hidden state at the time step that generates the parent of the current node in the partial tree; $\mathbf{s}^{\text{word}}_t$ is the hidden state of the most recent time step before $t$ that generated a subword (note that $\mathbf{s}^{\text{word}}_t$ comes from the word RNN, discussed below); and $[\cdot]$ denotes a concatenation.

Meanwhile, if at the previous time step $t-1$, \treedec~did not generate a CFG rule, then the update at time step $t$ must come from a subword being generated by the word RNN. In that case, we also update the rule RNN similarly by replacing the embedding of the CFG rule with the embedding of the subword. 

\paragraph{Word RNN.} At any time step $t$, if the word RNN is invoked, its hidden state $\mathbf{s}^\text{word}_t$ is:
\begin{equation*}
  \small
  \mathbf{s}^\text{word}_t = \text{LSTM}([\mathbf{s}^{\text{tree}}_p; \mathbf{w}_{t-1}; \mathbf{c}_{t-1}], \mathbf{s}^{\text{word}}_{t-1}),
\end{equation*}
where $\mathbf{s}^{\text{tree}}_p$ is the hidden state of rule RNN that generated the CFG rule above the current terminal; $\mathbf{w}_{t-1}$ is the embedding of the word generated at time step $t-1$;
and $\mathbf{c}_{t-1}$ is the attention context computed at the previous word RNN time step $t-1$.

\paragraph{Softmax.}~At any step $t$, our softmax logits are $\mathbf{W} \cdot \tanh{[\mathbf{s}^{\text{tree}}_t, \mathbf{s}^{\text{word}}_t]}$, where $\mathbf{W}$ varies depending on whether a rule or a subword unit is needed.

\begin{figure}[t!]
\begin{center}
  \includegraphics[width=0.48\textwidth]{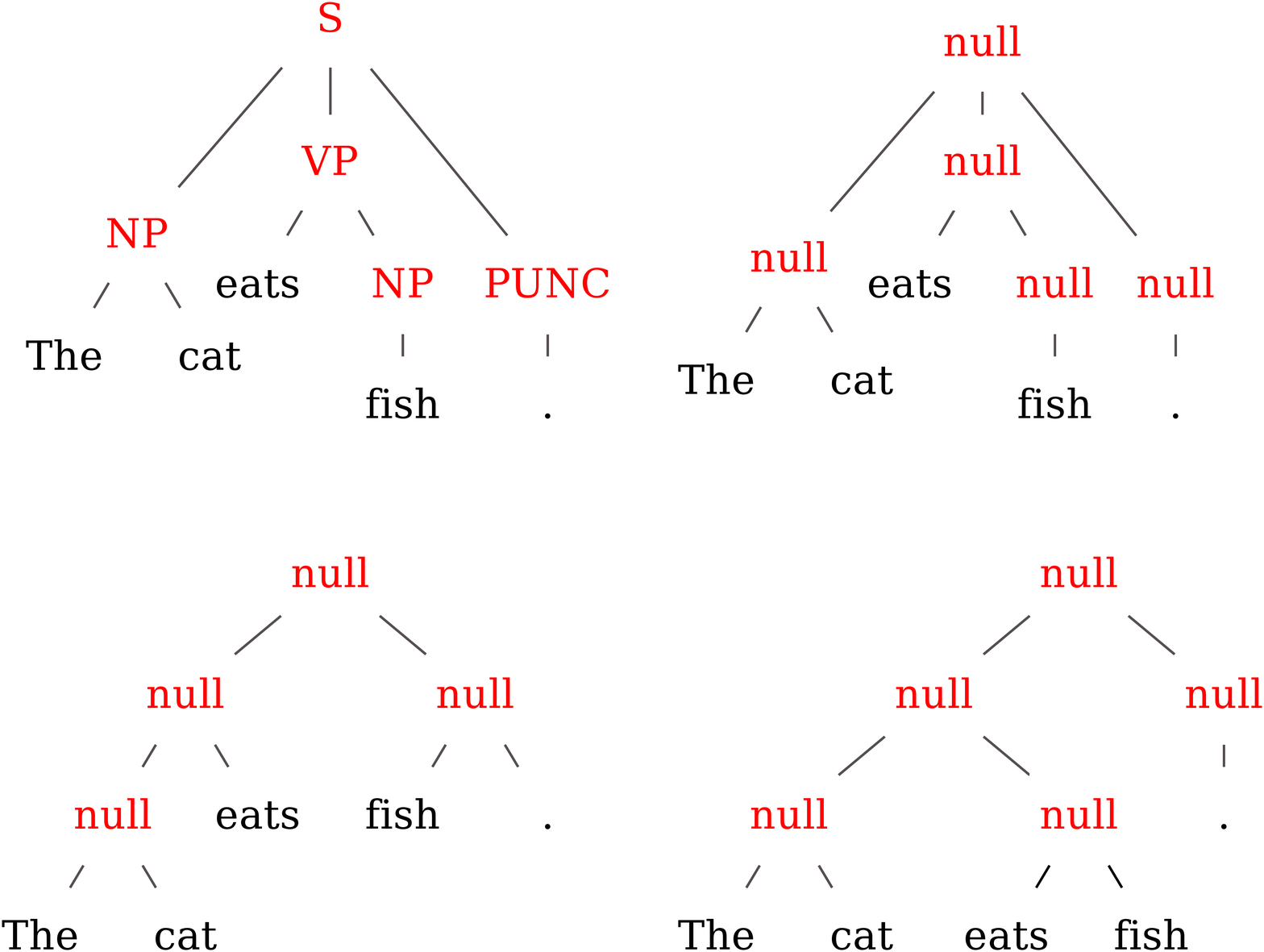}
  \captionof{figure}{\label{fig:trees}An example of four tree structures~(Details of preterminals and subword units omitted for illustration purpose).}
\end{center}
\end{figure}
\begin{figure}[t!]
\begin{center}
  \includegraphics[width=0.48\textwidth]{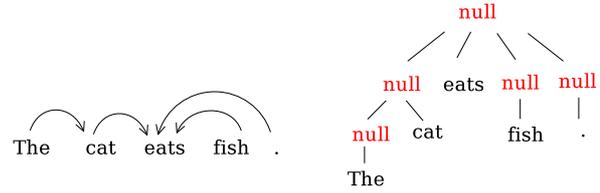}
  \captionof{figure}{\label{fig:dep_trees}Conversion of a dependency tree for \treedec. \textit{Left:} original dependency tree. \textit{Right:} after conversion.}
\end{center}
\end{figure}

\section{\label{sec:tree}Tree Structures}
Unlike prior work on syntactic decoders designed for utilizing a specific type of syntactic information~\citep{seq2dep}, \treedec~is a flexible NMT model that can utilize any tree structure. Here we consider two categories of tree structures:

\textbf{Syntactic Trees} are generated using a third-party parser, such as Berkeley parser~\citep{berkeley_parser_1,berkeley_parser_2}.
~\autoref{fig:trees}~\textit{Top Left} illustrates an example constituency parse tree. We also consider a variation of standard constituency parse trees where all of their nonterminal tags are replaced by a null tag, which is visualized in~\autoref{fig:trees}~\textit{Top Right}.
\\
In addition to constituency parse trees, \treedec~can also utilize dependency parse trees via a simple procedure that converts a dependency tree into a constituency tree. Specifically, this procedure creates a parent node with null tag for each word, and then attaches each word to the parent node of its head word while preserving the word order. An example of this procedure is provided in~\autoref{fig:dep_trees}.

\textbf{Balanced Binary Trees}
are syntax-free trees constructed without any linguistic guidance. We use two slightly different versions of binary trees. Version 1~(\autoref{fig:trees}~\textit{Bottom Left}) is constructed by recursively splitting the target sentence in half and creating left and right subtrees from the left and right halves of the sentence respectively. Version 2~(\autoref{fig:trees}~\textit{Bottom Right}), is constructed by applying Version 1 on a list of nodes where consecutive words are combined together. All tree nodes in both versions have the null tag. We discuss these construction processes in more detail in Appendix~\ref{sec:make_tree_alg}.

In the experiments detailed later, we evaluated \treedec~with four different settings of tree structures: 1) the fully syntactic constituency parse trees; 2) constituency parse trees with null tags; 3) dependency parse trees; 4) a concatenation of both version 1 and version 2 of the binary trees, (which effectively doubles the amount of the training data and leads to slight increases in accuracy).



\section{\label{sec:exp}Experiments}

\paragraph{Datasets.} We evaluate \treedec~on three datasets: 1) the KFTT (ja-en) dataset~\cite{kftt}, which consists of Japanese-English Wikipedia articles; 
2) the IWSLT2016 German-English (de-en) dataset~\cite{iwslt2016_eval}, which consists of TED Talks transcriptions; 
and 3) the LORELEI Oromo-English (or-en) dataset\footnote{LDC2017E29}, which largely consists of texts from the Bible. Details are in~\autoref{tab:dataset}. English sentences are parsed using Ckylark~\citep{ckylark} for the constituency parse trees, and Stanford Parser~\citep{stanford_parser,neural_dep_parser} for the dependency parse trees. We use byte-pair encoding~\citep{bpe} with 8K merge operations on ja-en, 4K merge operations on or-en, and 24K merge operations on de-en.

\begin{table}[h]
  \small
  \centering
  \begin{tabular}{l|ccc}
  \toprule
  \textbf{Dataset} & \textbf{Train} & \textbf{Dev} & \textbf{Test} \\
  \midrule
  ja-en & 405K & 1166 & 1160 \\
  de-en & 200K & 1024 & 1333 \\
  or-en     & 6.5K & 358  & 359 \\
  \bottomrule
  \end{tabular}
  \caption{\label{tab:dataset}\#~sentences in each dataset.}
\end{table}

\paragraph{Baselines.} We compare \treedec~against three baselines: 1) \seqtoseq: the standard \seqtoseq~model with attention; 2) CCG: a syntax-aware translation model that interleaves Combinatory Categorial Grammar (CCG) tags with words on the target side of a \seqtoseq ~model~\citep{CCG17}; 3) CCG-null: the same model with CCG, but all syntactic tags are replaced by a null tag; and 4) LIN: a standard \seqtoseq~model that generates linearized parse trees on the target side~\citep{str2lin_tree}.

\paragraph{Results.}  \autoref{tab:results} presents the performance of our model and the three baselines. For our model, we report the performance of \treedec-con, \treedec-con-null, \treedec-dep, and \treedec-binary~(settings 1,2,3,4 in \autoref{sec:tree}). On the low-resource or-en dataset, we observe a large variance with different random seeds, so we run each model with $6$ different seeds, and report the mean and standard deviation of these runs. \treedec-con-null and \treedec-con achieved comparable results, indicating that the syntactic labels have neither a large positive nor negative impact on \treedec. For ja-en and or-en, syntax-free \treedec~outperforms all baselines. On de-en, \treedec~loses to CCG-null, but the difference is not statistically significant~($p > 0.1$).
%

\begin{table}[h]
 \small
  \centering
  \begin{tabular}{l|cccc}
  \toprule
  \multirow{2}{*}{\textbf{Model}} &
  \multirow{2}{*}{\textbf{ja-en}} &
  \multirow{2}{*}{\textbf{de-en}} &
  \textbf{or-en} \\
  & & & ($\text{mean} \pm \text{std}$) \\
  \midrule
  \seqtoseq    & $21.10$  & $32.26$ & $10.90 \pm 0.57$ \\
  CCG          & $22.44$ & $32.84$ & $12.55 \pm 0.60$ \\
  CCG-null          & $21.31$ & $\mathbf{33.10}$ & $11.96 \pm 0.57$ \\
  LIN          & $21.55$ & $31.79$ & $12.66 \pm 0.61$ \\
  \midrule
  \treedec-con& $21.59$ & $31.93$ & $11.43 \pm 0.58$ \\
  \treedec-con-null& $22.72$ & $31.21$ & $11.35 \pm 0.55$ \\
  \treedec-dep& $21.41$ & $31.23$ & $8.40 \pm 0.5$ \\
  \treedec-binary& $\mathbf{23.14^{*}}$ & $32.65$ & $\mathbf{13.10^{**} \pm 0.61}$ \\
  \bottomrule
  \end{tabular}
  \caption{\label{tab:results}BLEU scores of \treedec~and other baselines. Statistical significance is indicated with $*$ ($p < 0.05$) and $**$ ($p < 0.001$), compared with the best baseline.}
\end{table}

\begin{figure}[ht]
\begin{center}
  \includegraphics[width=0.45\textwidth]{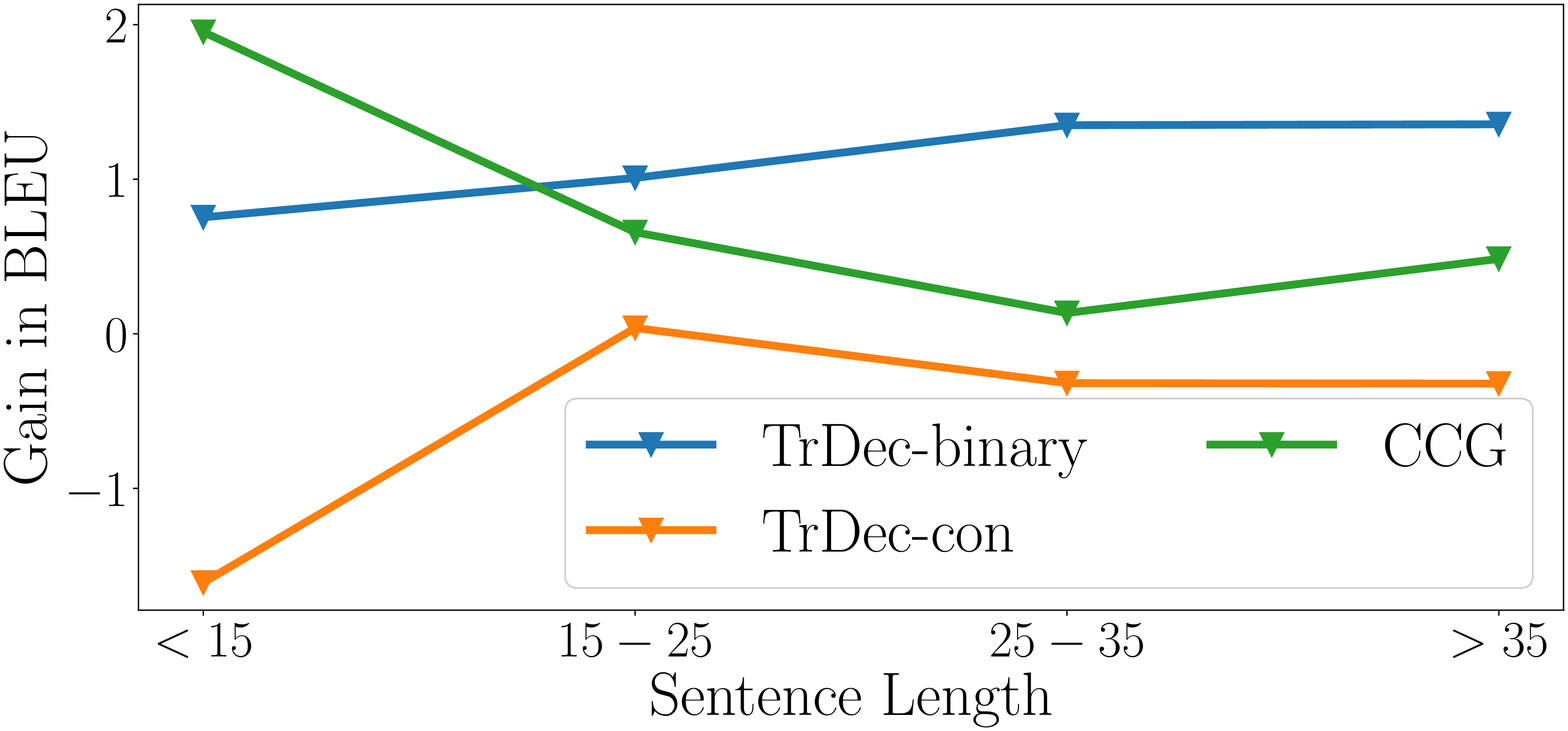}
  \captionof{figure}{\label{fig:gain_by_length}The gains of BLEU score over \seqtoseq.}
\end{center}

\begin{center}
  \includegraphics[width=0.45\textwidth]{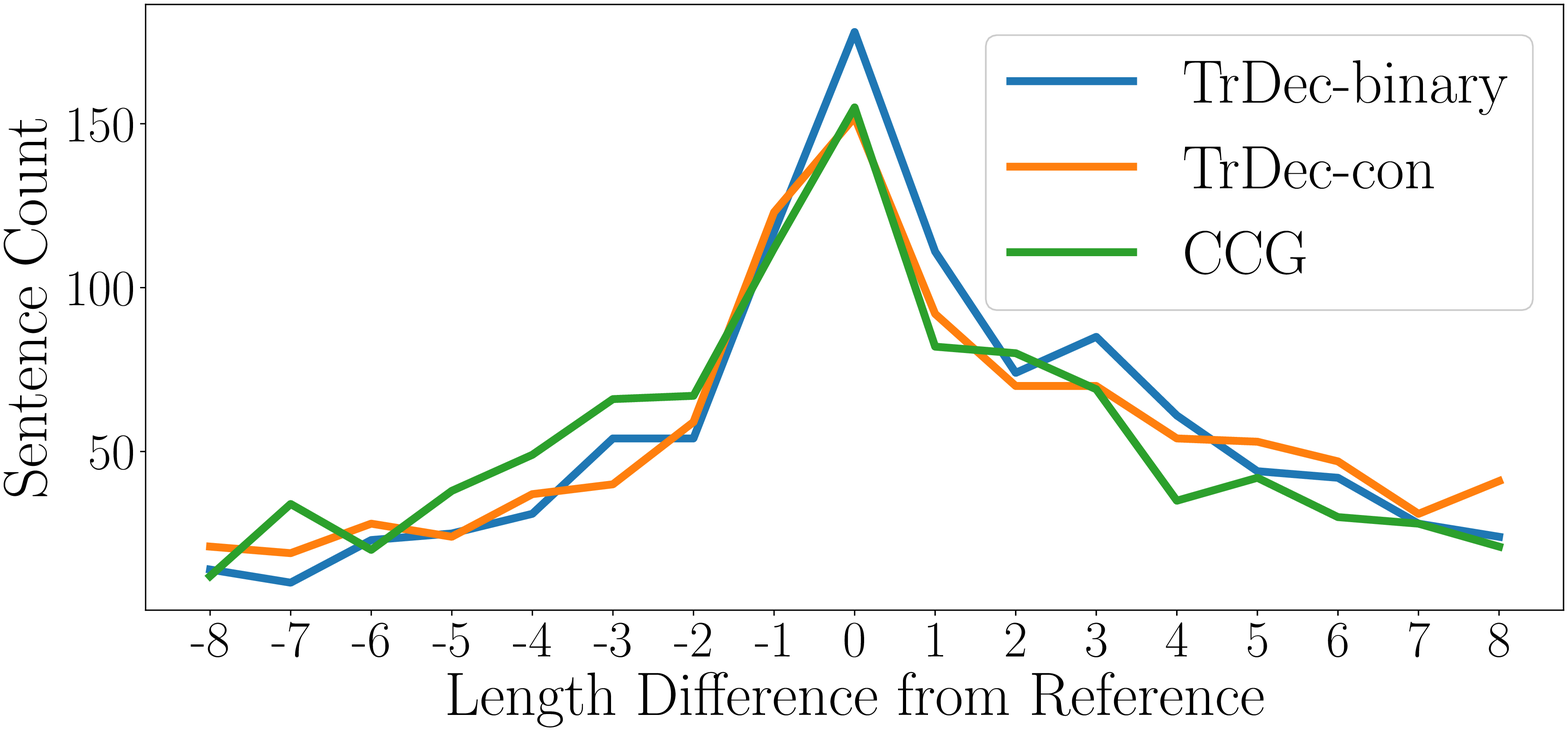}
  \captionof{figure}{\label{fig:length_count}Distribution of length difference from reference.}
\end{center}
\end{figure}

\paragraph{Length Analysis.} We performed a variety of analyses to elucidate the differences between the translations of different models, and the most conclusive results were through analysis based on the length of the translations. First, we categorize the ja-en test set into buckets by length of the reference sentences, and compare the models for each length category. \autoref{fig:gain_by_length} shows the gains in BLEU score over \seqtoseq~for the tree-based models. Since \treedec-con outperforms \treedec-dep for all datasets, we only focus on \treedec-con for analyzing \treedec's performance with syntactic trees. The relative performance of CCG decreases on long sentences. However, \treedec, with both parse trees and syntax-free binary trees, delivers more improvement on longer sentences. This indicates that \treedec~is better at capturing long-term dependencies during decoding. Surprisingly, \treedec-binary, which does not utilize any linguistic information, outperforms \treedec-con for all sentence length categories. 

Second, \autoref{fig:length_count} shows a histogram of translations by the length difference between the generated output and the reference. This provides an explanation of the difficulty of using parse trees. Ideally, this distribution will be focused around zero, indicating that the MT system is generating translations about the same length as the reference. However, the distribution of \treedec-con is more spread out than \treedec-binary, which indicates that it is more difficult for \treedec-con to generate sentences with appropriate target length. This is probably because constituency parse trees of sentences with similar number of words can have very different depth, and thus larger variance in the number of generation steps, likely making it difficult for the MT model to plan the sentence structure a-prior before actually generating the child sentences.

\section{\label{sec:conclusion}Conclusion}
We propose \treedec, a novel tree-based decoder for NMT, that generates translations along with the target side tree topology. We evaluate \treedec~on both linguistically-inspired parse trees and synthetic, syntax-free binary trees. Our model, when used with synthetic balanced binary trees, outperforms CCG, the existing state-of-the-art in incorporating syntax in NMT models.

The interesting result that syntax-free trees outperform their syntax-driven counterparts elicits a natural question for future work: how do we better model syntactic structure in these models? It would also be interesting to study the effect of using source-side syntax together with the target-side syntax supported by \treedec. 
\section*{Acknowledgements}
This material is based upon work supported in part by the Defense Advanced Research Projects Agency Information Innovation Office (I2O) Low Resource Languages for Emergent Incidents (LORELEI) program under Contract No. HR0011-15-C0114, and the National Science Foundation under Grant No. 1815287. The views and conclusions contained in this document are those of the authors and should not be interpreted as representing the official policies, either expressed or implied, of the U.S. Government. The U.S. Government is authorized to reproduce and distribute reprints for Government purposes notwithstanding any copyright notation here on.

\bibliography{main}
\bibliographystyle{acl_natbib_nourl}

\clearpage
\newpage

\appendix

\section{\label{sec:appendix}Appendix}
\subsection{\label{sec:make_tree_alg} Algorithm}
Here we list two simple algorithms for making balanced binary trees on the target sentence. For our experiments of \treedec~on binary trees, we use both algorithms to produce two versions of binary tree for each training sentence, and concatenate them as a form of data augmentation strategy.
%

\begin{algorithm}[h]
  \caption{The first method of making balanced binary tree}
  \small
  \DontPrintSemicolon
  \SetKwInOut{Input}{Input}
  \SetCommentSty{itshape}
  \SetKwComment{Comment}{$\triangleright$\ }{}
  \SetKwInOut{Output}{Output}
  \SetKwFunction{FMakeTree}{make\_tree\_v1}
  \SetKwProg{Fn}{Function}{:}{}
  \Input{$w$: the list of words in a sentence, $l$: start index, $r$: end index}
  \Output{a balanced binary tree for words from $l$ to $r$ in $w$}
  \Fn{\FMakeTree{$w$, $l$, $r$}}{
    \If {$l = r$} { 
      \KwRet $\textrm{TerminalNode}(w[l])$\;
    }
    \;
    $m = \text{floor}((l + r) / 2)$ \Comment*[r]{index of split point} 
    $\textit{left\_tree} = \textrm{make\_tree\_v1}(w, l, m)$ \;
    $\textit{right\_tree}= \textrm{make\_tree\_v1}(w, m+1, r)$ \;
    \;
    \KwRet $\textrm{NonTerminalNode}(\textit{left\_tree}, \textit{right\_tree})$
  }
\end{algorithm}

\begin{algorithm}[h]
  \caption{The second method of making balanced binary tree}
  \small
  \DontPrintSemicolon
  \SetKwInOut{Input}{Input}
  \SetCommentSty{itshape}
  \SetKwComment{Comment}{$\triangleright$\ }{}
  \SetKwInOut{Output}{Output}
  \SetKwFunction{FMakeTree}{make\_tree\_v2}
  \SetKwProg{Fn}{Function}{:}{}
  \Input{$w$: the list of words in a sentence, $l$: start index, $r$: end index (inclusive)}
  \Output{a balanced binary tree for words from $l$ to $r$ in $w$}
  \Fn{\FMakeTree{$w$, $l$, $r$}}{
    $\textit{nodes} = \textrm{EmptyList}()$ \;
    $i = 0$ \;

    \While{$i < \text{len}(w)-1$} {
       $\textit{lc} = \textrm{TerminalNode}(w[i])$ \;
       $\textit{rc} = \textrm{TerminalNode}(w[i+1])$ \;
       $\textit{n} = \textrm{NonTerminalNode}(\textit{lc}, \textit{rc})$ \;
       $\textit{nodes}.\textrm{append}(n)$ \;
       $i = i + 2$ \;
    }
    \;
    \If{$i \neq \textit{len}(w)$} {
      $n = \textrm{TerminalNode}(w[i])$ \;
      $\textit{nodes}.\textrm{append}(n)$ \;
    }
    \;

    \KwRet $\text{make\_tree\_v1}(\textit{nodes}, 0, \textit{len}(\textit{nodes}) - 1)$ \;
  }
\end{algorithm}
\end{document}